\documentclass[11pt]{article}

\usepackage[numbers]{natbib}

\usepackage{booktabs}       
\usepackage{amsfonts}       
\usepackage{amsmath}
\usepackage{graphicx}
\usepackage{amsthm}
\usepackage{amssymb}
\usepackage{multirow}
\usepackage{makecell}
\usepackage[vlined,linesnumbered,ruled,resetcount]{algorithm2e}
\usepackage[colorlinks,linkcolor=magenta,filecolor=blue,citecolor=blue,urlcolor=blue]{hyperref}%
\usepackage[top=1in, left=1in, right=1in, bottom=1in]{geometry}
\usepackage{subfigure}
\usepackage{amssymb}
\usepackage{pifont}
\usepackage{mathtools}
\usepackage{stmaryrd}
\usepackage[T1]{fontenc}

\usepackage{authblk}
\usepackage{algorithmic}
\usepackage[inline]{enumitem}
\usepackage{subcaption}
\usepackage{xcolor}
\usepackage{colortbl}

\DeclareMathOperator*{\argmin}{arg\,min}
\DeclareMathOperator*{\CQ}{cq}

\newcommand{\cmark}{\ding{51}}
\newcommand{\xmark}{\ding{55}}

\newcommand{\greyrule}{\arrayrulecolor{black!10}\midrule\arrayrulecolor{black}}

\theoremstyle{definition}

\newcommand{\wh}{\widehat}

\renewcommand{\epsilon}{\varepsilon}
\renewcommand{\phi}{\varphi}

\renewcommand{\hat}{\wh}

\makeatletter
\newcommand*{\RN}[1]{\expandafter\@slowromancap\romannumeral #1@}
\makeatother

\makeatletter
\newcommand{\printfnsymbol}[1]{%
  \textsuperscript{\@fnsymbol{#1}}%
}
\makeatother

\title{KV Cache is 1 Bit Per Channel: Efficient Large Language Model Inference with Coupled Quantization}

\author[1]{Tianyi Zhang}
\author[1]{Jonah Yi}
\author[2]{Zhaozhuo Xu}
\author[1,3]{Anshumali Shrivastava}
\affil[1]{Department of Computer Science, Rice University}
\affil[2]{Department of Computer Science, Stevens Institute of Technology}
\affil[3]{ThirdAI Corp.}
\affil[ ]{{\texttt{\{tz21, jwy4, anshumali\}@rice.edu}, \texttt{zxu79@stevens.edu}}}
\begin{document}

\maketitle
\begin{abstract}
    Efficient deployment of Large Language Models (LLMs) requires batching multiple requests together to improve throughput. As the batch size, context length, or model size increases, the size of the key and value (KV) cache can quickly become the main contributor to GPU memory usage and the bottleneck of inference latency. Quantization has emerged as an effective technique for KV cache compression, but existing methods still fail at very low bit widths. We observe that distinct channels of a key/value activation embedding are highly inter-dependent, and the joint entropy of multiple channels grows at a slower rate than the sum of their marginal entropies. Based on this insight, we propose Coupled Quantization (CQ), which couples multiple key/value channels together to exploit their inter-dependency and encode the activations in a more information-efficient manner. Extensive experiments reveal that CQ outperforms or is competitive with existing baselines in preserving model quality. Furthermore, we demonstrate that CQ can preserve model quality with KV cache quantized down to 1-bit.
\end{abstract}

\section{Introduction}

Large Language Models (LLMs) have showcased remarkable generalization abilities across various tasks, including text generation, language translation, and reasoning, without needing specific finetuning \citep{multitask_learners}. These impressive capabilities have empowered LLMs to find applications in numerous domains, such as law \citep{apps_of_llm}, education \citep{edu_llm}, and patient care \citep{health_llm}. However, the high computational demands and the prohibitive deployment costs of LLMs have created significant barriers, hindering their widespread adoption \citep{apps_of_llm, frugalgpt}. Particularly, as LLMs move towards larger model size \citep{compute_optimal_llm} and longer context length \citep{long_context_llm}, they require faster graphics processing units (GPUs), or other specialized processors, with higher memory capacity for efficient inference. Hence it is crucial to develop approaches for reducing the computational costs and memory requirement of LLMs.

To accelerate LLM inference, key and value (KV) caching \citep{scissorhands} has been proven to be an effective technique without affecting model quality. In autoregressive decoder-only LLMs, KV caching works through trading off memory for computations: the key and value activations of all previous tokens in the current batch are saved in memory to avoid their recomputation for generating the next token. However, since KV cache scales linearly with the number of tokens and batch size, it can quickly overwhelm the memory capacity of existing GPUs under long context or large batch size settings. In addition, since past key and value activations are not shared between sequences (except for maybe a common prompt), reading the KV cache from GPU memory becomes the primary inference bottleneck as opposed to the computation of the attention scores and value activations of the next token \citep{kvquant}. Thus, it is worthwhile to explore techniques for compressing KV cache for two primary benefits: \begin{enumerate*}
    \item speeding up LLM inference through reducing the amount of memory reads for KV cache,
    \item lowering the GPU memory requirements of inference for a given batch size and context length.
\end{enumerate*} Existing approaches typically compress KV cache through token eviction \citep{h2o,scissorhands} or activation quantization \citep{kivi, kvquant}. While they can preserve model quality at moderate compression rates (4$\times$ compression or 4 bits per floating-point number), model quality quickly deteriorates at high compression rates (16$\times$ compression or 1 bit per floating-point number). In this work, we propose Coupled Quantization (CQ), a novel KV cache quantization method that preserves model quality up to 16$\times$ compression or 1 bit per floating-point number.

Our approach is motivated by the observation that different channels within the same key/value activation embedding are highly inter-dependent. Thus, it is more information efficient to encode multiple channels of a key/value activation embedding than quantizing each channel independently. Existing KV cache quantization methods employ per-channel or per-token quantization strategies, which fail to exploit the inter-dependence between channels and suffer catastrophic model quality degradation at high compression rates. Our proposed method exploits the mutual dependency between channels by jointly quantizing multiple channels and achieves better preservation of model quality than existing approaches in most cases, especially under low bit width settings. We summarize our contributions as follows,
\begin{enumerate}[itemsep=0mm]
    \item We empirically observe the phenomenon that different channels within the same key/value activation embedding in an LLM share a high amount of dependency or mutual information, which is a key insight not leveraged by existing KV cache compression approaches.
    \item We propose Coupled Quantization (CQ), a novel KV cache quantization method that takes advantage of the reduced entropy of jointly encoding multiple channels.
    \item Through extensive experiments, we demonstrate the effectiveness of CQ against the most competitive existing methods. Furthermore, we showcase the ability of CQ in preserving model quality at an extreme KV cache compression level of 1-bit.
\end{enumerate}

\section{Background}

In this section, we introduce the relevant background and context including the KV caching technique, the von Neumann bottleneck of KV cache, and channel-wise quantization.

\subsection{LLM Attention and KV Cache}

Decoder-only transformer-based LLMs employ masked self-attention \citep{attention}, in which activations of the current token is only dependent on previous tokens and unaffected by future ones. This property enables training parallelism for the next-token prediction objective, and gives rise to the KV caching technique for efficient inference decoding. Consider the decoding step for the $t$-th token in a single head of attention in an LLM. The input embedding of the $t$-th token (a column vector), $e_t$, goes through three distinct transformations to become key, query, and value activation embeddings $f_K(e_t), f_Q(e_t), f_V(e_t)$, where the transformations $f_K, f_Q, f_V$ are composed of linear projection and positional encoding such as RoPE \citep{rope}. The output embedding of attention for the $t$-th token is computed as
\begin{equation}
\label{eq:attn}
    \mathrm{attention}(e_t) = \biggl[ \begin{array}{ccc}
        f_V(e_1) & \dots & f_V(e_t)
  \end{array} \biggr] \mathrm{softmax}\Biggl(\biggl[ \begin{array}{ccc}
        f_K(e_1) & \dots & f_K(e_t)
  \end{array} \biggr]^\top f_Q(e_t)\Biggr)
\end{equation}
Thus, computing the output embedding of the current token requires the key and value activation embeddings of all previous tokens, $f_K(e_i)$ and $f_V(e_i)$ where $i \in \left\{1, \dots, t-1\right\}$. These embeddings are cached in memory from previous decoding steps as \textit{KV cache} to avoid redundant computations and reduce inference latency. The size of KV cache can be calculated as $b \times n \times l \times 2 \times h \times c$ floating-point numbers, where $b$ is the batch size, $n$ is the number of tokens in each sequence, $l$ is the number of layers in the model, $2$ is for key and value, $h$ is the number of key/value attention heads, and $c$ is the number of channels in a single head of key/value activation embedding. As batch size, context length, or model size increases, the size of KV quickly overwhelms limited GPU memory.

\subsection{The von Neumann Bottleneck of KV Cache}

The attention computation in Equation \ref{eq:attn} is primarily bottlenecked by GPU memory bandwidth, known as the von Neumann bottleneck, due to low compute-to-global-memory-access ratio. GPUs are able to perform computations significantly faster than reading from global memory, and previous works have shown that attention computation can be accelerated by avoiding unnecessary global memory reads/writes through kernel fusion \citep{flashattention}. During the decoding phase in LLM attention, computing the output for the current token requires fetching the cached key and value embeddings of all previous tokens from global memory, resulting in a low ratio of arithmetic operations to global memory reads \citep{kvquant}. Furthermore, each sequence in a batch retains its own KV cache, leading to poor utilization of the parallel processing capabilities of GPUs. Therefore, KV cache compression algorithms can mitigate the von Neumann bottleneck and improve LLM inference efficiency, even if the approach introduces additional computational overheads in decompression or dequantization. As GPU compute cores are mostly stalled by KV cache memory accesses in attention, quantization approaches can effectively reduce memory reads while introducing negligible latency from dequantization computations.

\subsection{Channel-wise Quantization}

Existing KV cache quantization methods \citep{kvquant,kivi} employ channel-wise quantization for keys and token-wise quantization for values, based on the observation that certain key channels exhibit outlier patterns in magnitude while value channels do not. Channel-wise and token-wise quantization are similar, except the direction along which the quantization centroids are learned. In non-uniform channel-wise quantization, a set of centroids is learned for each channel. Suppose $A$ is a key or value activation matrix, and $A_{i,*}$ denotes the $i$-th channel of $A$. Then, non-uniform $b$-bit channel-wise quantization aims to learn a set of centroids $C_i^\star \subset \mathbb R$ for each channel $i$ of $A$ independently through the objective
\begin{equation}
    C_i^\star = \argmin_{\substack{C \subset \mathbb R \\ |C|=2^b}} \Big\lVert A_{i,*} - q(A_{i,*})\Big\rVert_2^2
\end{equation}
where $q$ quantizes each value in $A_{i,*}$ to the nearest centroid in $C$.
\section{Methodology}

In this section, we motivate our proposal using information theory and introduce the Coupled Quantization (CQ) approach for KV cache compression.

\begin{figure}[t]
  \centering
  \includegraphics[width=0.8\textwidth]{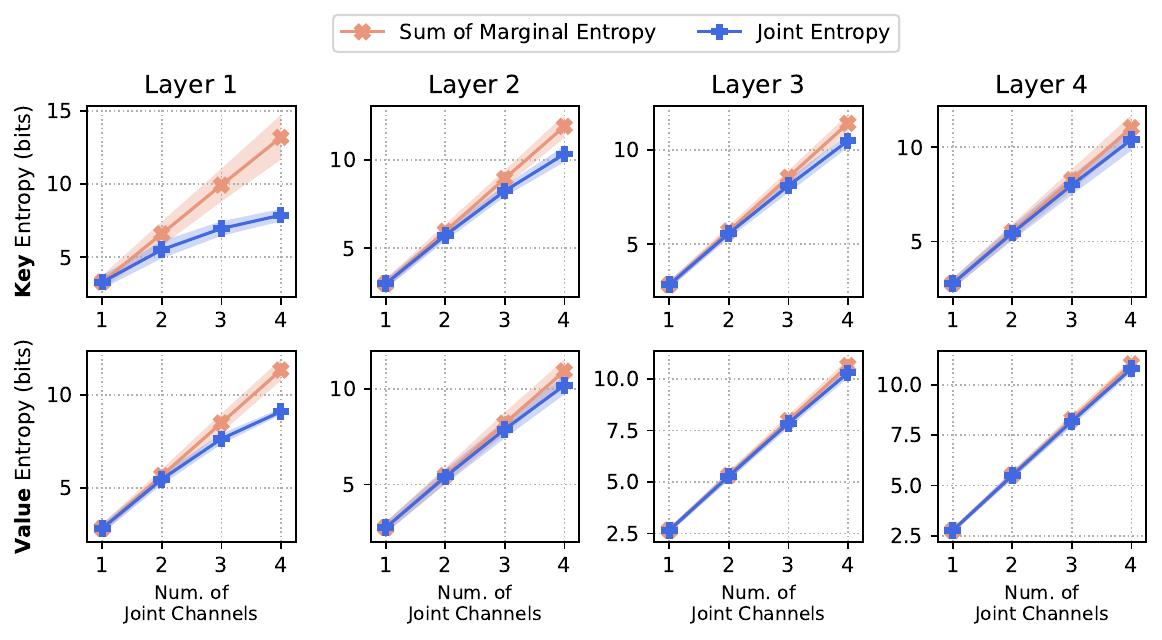}
  \caption{Growth rate of joint entropy versus sum of marginal entropies of the LLaMA-7b key/value activation embeddings on 262k tokens of WikiText-2. Entropy is estimated using Equation \ref{eq:entropy}. The slower growth rate of joint entropy implies that jointly quantizing more channels requires fewer bits than quantizing each channel independently.}
  \label{fig:entropy}
\end{figure}

\begin{figure}[t]
  \centering
  \includegraphics[width=\textwidth]{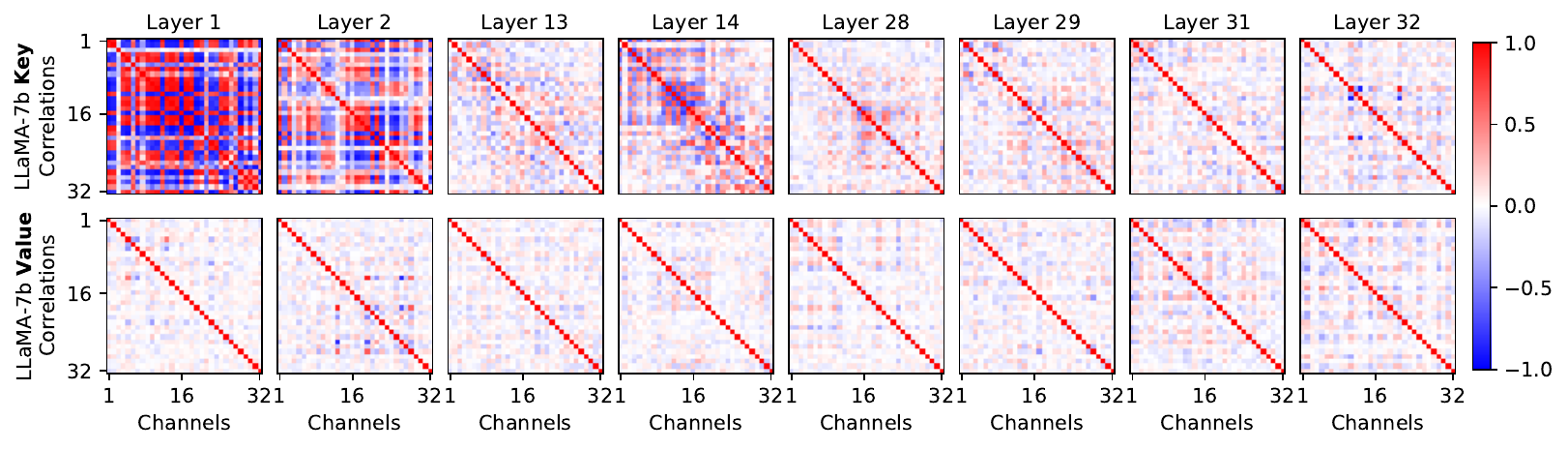}
 \caption{Correlation matrices of the first 32 channels of 8 layers of LLaMA-7b key and value activation embeddings on WikiText-2. Channel pairs exhibit high levels of linear dependency, shown by high magnitudes of the correlation coefficients.}
  \label{fig:corr}
\end{figure}

\subsection{Motivations}

Our proposed approach is inspired by concepts in information theory \citep{information_theory}. We consider each channel in a key/value activation embedding as a random variable $X_1, X_2, \dots$. The amount of information (or uncertainty) in channel $X$ can be measured by \textit{entropy}, defined as $H(X) = -\int_{\mathbb X} p(x) \log_2 p(x) \,dx$, where $p(\cdot)$ is the probability density function and $\mathbb X$ is the support of $X$. $H(X)$ measures the theoretical number of bits needed for losslessly encoding the channel $X$, so it can be used to gauge how ``quantizable'' a channel is: if $H(X_1) < H(X_2)$, then channel $X_1$ may be quantized to fewer bits than channel $X_2$ while achieving the same quantization error.

Our insight is that different channels from the same key/value activation embedding may be interdependent, which would reduce the number of bits required for jointly encoding multiple channels together compared to encoding them independently. The total amount of information (or uncertainty) in two channels $X_1, X_2$ is measured by \textit{joint entropy}, defined as $H(X_1, X_2) = -\int_{\mathbb X_1}\int_{\mathbb X_2} p(x_1, x_2) \log_2 p(x_1, x_2) \,dx_2\,dx_1$, where $p(\cdot, \cdot)$ is the joint probability density function. The joint entropy of two channels is the difference between the sum of their marginal entropies and their mutual information, i.e., $H(X_1, X_2) = H(X_1) + H(X_2) - I(X_1, X_2)$, where $I(\cdot, \cdot)$ is a non-negative quantity for measuring the mutual dependency of two random variables. Thus, we have 
\begin{equation}
    H(X_1, X_2) \le H(X_1) + H(X_2)
\end{equation}
which implies the number of bits needed for jointly encoding two channels is no more than the total number of bits needed for encoding them independently. Previous works have demonstrated that deep neural networks \citep{low_rank_deep_net} and attention-based networks \citep{attention_loses_rank} tend to produce low-rank embeddings, which suggests that channels of key/value embedding in LLM may exhibit high amount of mutual dependency.

It is hence beneficial to measure the difference between the joint entropy of multiple channels and the sum of their marginal entropies in key and value activation embeddings. A significant difference would suggest that encoding these channels together is more information-efficient than encoding them independently. However, it is intractable to derive the exact entropy or joint entropy of channels, since their probability density functions are not known. Therefore, we employ the ``binning'' trick \citep{estimating_mi} to estimate entropy. We first observe an empirical distribution of key and value channels by saving the KV cache on a dataset, and partition the support of each channel into equally sized bins. Then, values of each channel are discretized to the index of the bin they fall into. Finally, the joint entropy of $n$ channels $X_1, \dots, X_n$ is estimated with the Riemann sum,
\begin{equation}
\label{eq:entropy}
    H(X_1, \dots, X_n) \approx \sum_{x_1 \in \mathbb B_1} \dots \sum_{x_n \in \mathbb B_n} \hat p(x_1, \dots, x_n) \log_2 \hat p(x_1, \dots, x_n)
\end{equation}
where $\mathbb B_i$ is the support of the binned or discretized $X_i$ and $\hat p(\cdot)$ is the empirical probability density function. Specifically, we divide the channels of key and value embeddings of LLaMA-7b into non-overlapping groups of $c$ contiguous channels, where $c \in \{1, 2, 3, 4\}$, and estimate the joint entropy and the sum of marginal entropies of each group. The support of each channel is partitioned into 16 equally sized bins. Figure \ref{fig:entropy} shows the mean and standard deviation of the estimated joint entropy and sum of marginal entropies of three layers of LLaMA-7b on 262k tokens of the WikiText-2 dataset, averaged over groups. We only show a maximum group size of 4, since increasing the group size requires saving exponentially more key and value embeddings to avoid empty bins and maintain estimation quality. As shown in Figure \ref{fig:entropy}, the sum of marginal entropies grows at a linear rate while the joint entropy increases slower at a sub-linear rate. This implies that as the number of jointly quantized channels increases, the total amount of information needed for encoding decreases. This phenomenon is the foundation that motivates our proposed approach.

In addition to presenting the estimated marginal and joint entropy, we also show the Pearson correlation coefficients between channels of LLaMA-7b key and value activation embeddings on WikiText-2. Correlation coefficient captures the linear dependency between two random variables. Heat maps for the correlation matrices of 8 layers are shown in Figure \ref{fig:corr}, while the correlation matrices of all layers are presented in Section \ref{sec:corr} of the Appendix. The key and value channels exhibit high levels of linear dependency and they are clearly not independently distributed, as shown by high magnitudes of the correlation coefficients.

\begin{figure}[t]
  \centering
  \includegraphics[width=\textwidth]{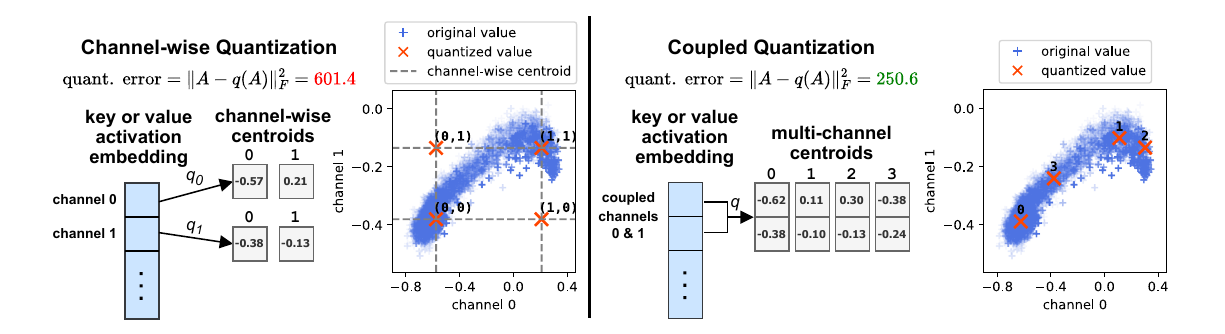}
  \caption{A comparison of 1-bit channel-wise quantization and our proposed Coupled Quantization (using 2 bits per 2 channels as an example). The quantization results on the first two channels of the first-layer key activation embeddings of LLaMA-7b on the WikiText-2 dataset are shown. Channel-wise quantization is ineffective at capturing the original values at low widths, while CQ leverages the dependency between channels to achieve low quantization errors.}
  \label{fig:cq}
\end{figure}

\subsection{Coupled Quantization}

Motivated by the finding that distinct key/value channels exhibit high dependency, we propose Coupled Quantization (CQ), an information-efficient quantization approach for compressing LLM KV cache. Unlike existing KV cache quantization methods which quantizes channel-wise or token-wise, CQ performs channel-coupled quantization for keys and values. More concretely, channels of a key or value activation embedding are divided into equally sized, non-overlapping groups of contiguous channels. The channels in each group are \textit{coupled}, as they are jointly quantized and share a single quantization code. For each group of coupled channels, a distinct set of multi-channel centroids are learned, where each centroid has dimensionality equal to the number of channels in that group. When quantizing a key or value activation embedding, each group of coupled channels are quantized to the nearest centroid in terms of L2 distance. We use the CQ-\texttt{<c>}c\texttt{<b>}b notation to denote the configuration of channel coupling and quantization bit width, where \texttt{<c>} is the number of channels in each group and \texttt{<b>} indicates the number of bits in a quantized code for a group. For example, CQ-4c8b means that every 4 contiguous channels are coupled together and each coupled group shares an 8-bit code, which is equivalent to 2-bit channel-wise quantization in terms of storage overhead of quantized codes. A illustrative comparison of channel-wise quantization and CQ is shown in Figure \ref{fig:cq}. Although previous works \citep{kvquant,kivi} opt to quantize keys channel-wise and values token-wise, we adopt channel-coupled quantization for both keys and values. Similar to existing approaches \citep{kvquant,kivi}, CQ quantizes keys before RoPE \citep{rope} is applied, which increases the quantization difficulty by introducing more outliers in key activations.

\begin{figure}[t]
  \centering
  \includegraphics[width=\textwidth]{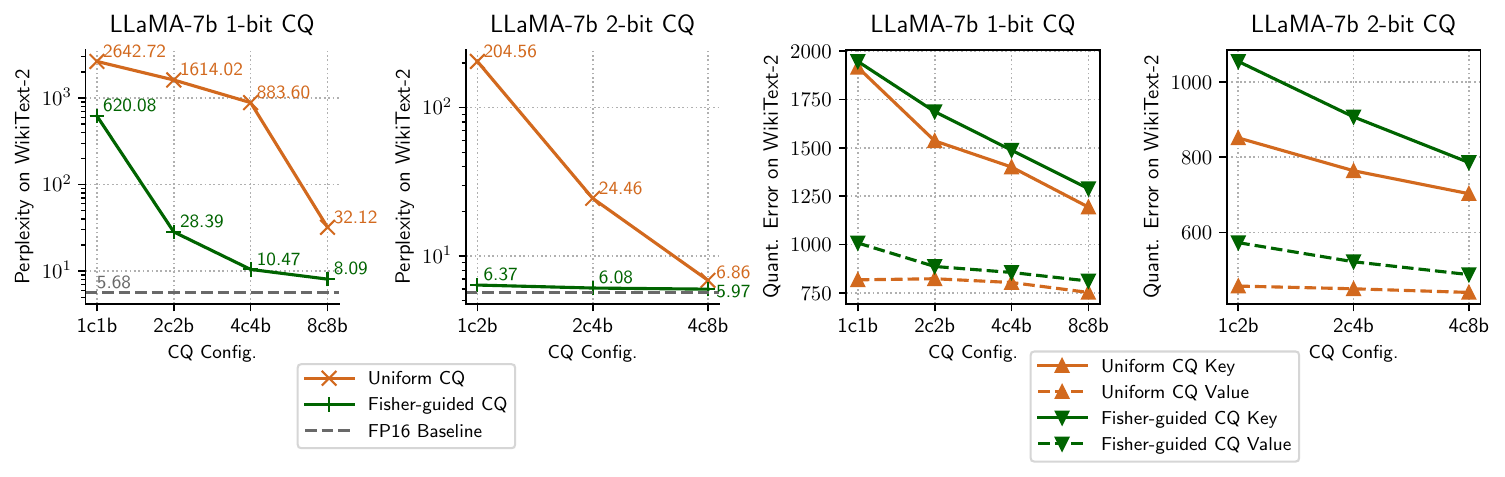}
  \caption{Perplexity and key/value quantization errors (averaged over all layers) of LLaMA-7b on WikiText-2. Channels coupling and Fisher-guided centroid learning are effective for improving perplexity.}
  \label{fig:ppl}
\end{figure}

\subsubsection{Centroid Learning}

In CQ, the multi-channel centroids for each channel group are learned offline on a calibration dataset by leveraging uniform clustering or second-order-information-informed clustering. Specifically, for uniform centroid learning of the CQ-$\,c\,$c$\,b\,$b configuration, a set of centroids $C_i^\star \subset \mathbb R^c$ is learned independently for each channel group $i$ through the objective
\begin{equation}
    C_i^\star = \argmin_{\substack{C \in \mathbb R^c\\ \left|C\right|=2^b}}\Big\lVert A_{ic:(ic+c-1),\,*} - \CQ\big(A_{ic:(ic+c-1),\,*}\big) \Big\rVert^2_F
\end{equation}
where $A_{ic:(ic+c-1),\,*}$ is the sub-matrix of $A$ containing all coupled channels of the $i$-th group, and $\CQ$ quantizes each column vector to the nearest centroid in $C$ in terms of L2 distance. We use the k-means algorithm \citep{kmeans} with k-means++ initialization \citep{kmeanspp} to optimize the objective.

LLMs are more sensitive to the quantization precision of certain weights than others \citep{squeezellm}. To better preserve model quality, centroids of CQ should be learned to be biased towards preserving the precision of more important activations. To this end, we leverage an approximation to the Hessian to perform second-order-information-informed centroid learning. More concretely, we use the diagonals of the Fisher information matrix $\mathcal F$ to identify the more influential key/value activations and guide the centroid learning process. This method was proposed by \citet{brecq} and used by \citet{squeezellm} for channel-wise weight quantization, and we extend it to multi-channel CQ. For performing Fisher-guided centroid learning, we first save a key/value activation matrix $A$ and its gradient $g(A) = \frac{\partial}{\partial A} \mathcal L(A)$ on a calibration dataset, where $\mathcal L$ is the training loss function. We approximate the Hessian matrix using the diagonals of the Fisher information matrix, which is the element-wise square of the gradient matrix  $\mathrm{diag}(\mathcal F) = g(A) \odot g(A)$. We use the sum of diagonal entries of the Fisher information matrix as a measure of importance for each group of channel-coupled activations, and obtain the centroid set $C_i^\star$ for the $i$-th channel group using the objective
\begin{equation}
    C_i^\star = \argmin_{\substack{C \subset \mathbb R^c\\ \left|C\right|=2^b}}\sum_j g\big(A_{ic:(ic+c-1),\,j}\big)^\top g\big(A_{ic:(ic+c-1),\,j}\big)\Big\lVert A_{ic:(ic+c-1),\,j} - \CQ(A_{ic:(ic+c-1),\,j}) \Big\rVert^2_F
\end{equation}
We leverage weighted k-means to optimize the objective. The overhead of the centroid learning process and centroid storage are discussed in Section \ref{sec:overhead}.

We validate the effectiveness of our proposed channel-coupling and Fisher-guided centroid learning by compressing LLaMA-7b KV cache to 1-bit and 2-bit, and present the perplexity results and quantization errors ($ \lVert A - \CQ(A)\rVert^2_F $ averaged over layers) on WikiText-2 under different CQ configurations in Figure \ref{fig:ppl}. The experimental setup is given in Section \ref{sec:experiments}. As the number of coupled channels increases, perplexity and quantization errors improve significantly, approaching the FP16 baseline performance. Although Fisher-guided centroid learning increases the quantization error, it better preserves the salient activations and achieve lower perplexity.

\begin{table}[t]
\caption{Perplexity on WikiText-2 under different KV cache quantization methods at varying bit-width. The results of INT, NF, and KVQuant (excluding 1b and 1b-1\%) are from \citep{kvquant}. ``NaN'' means Not a Number, which is caused by numerical stability issues of quantization. Our method CQ consistently outperforms non-dense-and-sparse quantization methods, and performs better or on par with the dense-and-sparse method KVQuant-\texttt{<b>}b-1\%.}
\setlength\tabcolsep{4pt}
\centering
\small
\label{tab:ppl_wiki}
\begin{tabular}{lr|ccccc}
\toprule
 & Bits Per FPN & LLaMA-7b & LLaMA-13b & LLaMA-2-7b & LLaMA-2-13b & Mistral-7b \\
 \midrule
 \midrule
FP16 & 16 & 5.68 & 5.09 & 5.12 & 4.57 & 4.76 \\
 \midrule
INT4 & 4.00-4.01 & 5.98 & 5.32 & 5.66 & 5.01 & 4.97 \\
INT4-gs128 & 4.16 & 5.77 & 5.16 & 5.32 & 4.71 & 4.82 \\
NF4 & 4.00-4.01 & 5.87 & 5.23 & 5.47 & 4.90 & 4.91 \\
NF4-gs128 & 4.16 & 5.77 & 5.17 & 5.30 & 4.71 & 4.83 \\
KVQuant-4b & 4.00-4.02 & \underline{5.73} & \underline{5.15} & \underline{5.18} & \underline{4.63} & 4.81 \\
KVQuant-4b-1\% & 4.32-4.35 & \textbf{5.70} & \textbf{5.11} & \textbf{5.14} & \textbf{4.59} & \textbf{4.78} \\
CQ-2c8b & 4.00 & \textbf{5.70} & \textbf{5.11} & \textbf{5.14} & \textbf{4.59} & \underline{4.79} \\
 \midrule
INT2 & 2.00-2.01 & 11779 & 69965 & 4708 & 3942 & 573 \\
INT2-gs128 & 2.14 & 37.37 & 41.77 & 117.88 & 93.09 & 51.96 \\
NF2 & 2.00-2.02 & 3210.5 & 5785.6 & 13601 & 4035.6 & 902.51 \\
NF2-gs128 & 2.14 & 351.23 & 141.19 & 634.59 & 642.44 & 252.85 \\
KVQuant-2b & 2.00-2.02 & 8.17 & 7.29 & 9.75 & 29.25 & 7.33 \\
KVQuant-2b-1\% & 2.32-2.35 & \underline{6.06} & \underline{5.40} & \underline{5.50} & \underline{4.92} & \underline{5.16} \\
CQ-4c8b & 2.00 & \textbf{5.97} & \textbf{5.32} & \textbf{5.42} & \textbf{4.81} & \textbf{5.11} \\
 \midrule
KVQuant-1b & 1.00-1.02 & 321.58 & 1617.40 & NaN & 4709.83 & 203.73 \\
KVQuant-1b-1\% & 1.32-1.35 & 9.93 & 7.97 & 9.50 & 13.76 & 10.07 \\
CQ-8c8b & 1.00 & \underline{8.09} & \underline{7.02} & \underline{7.75} & \underline{6.55} & \underline{7.25} \\
CQ-8c10b & 1.25 & \textbf{6.78} & \textbf{6.00} & \textbf{6.25} & \textbf{5.47} & \textbf{5.90} \\
 \bottomrule
\end{tabular}
\end{table}

\section{Experiments}
\label{sec:experiments}
In this section, we perform extensive experiments to validate the effectiveness of our proposed CQ approach for KV cache compression. We first introduce the experimental setups including hardware, software, metrics, datasets, and baselines used. Then, we present the detailed empirical results and provide discussions. Finally, we perform an ablation study to validate the effectiveness of each component of our proposal.

\textbf{Hardware Testbed} Experiments are performed on a Linux server running Ubuntu 20.04, equipped with 2 AMD EPYC 7742 CPUs, 1.5TB RAM, and 4 NVIDIA A100 40GB GPUs.

\textbf{Software Implementation} Our software implementation of CQ is based on PyTorch \citep{pytorch} and the HuggingFace Transformers library \citep{huggingface}.

\textbf{Evaluation Metrics and Benchmarks} We evaluate the quality of 5 popular open-source LLMs on various benchmarks under different KV cache quantization algorithms. The 5 LLMs considered are \begin{enumerate*}
    \item LLaMA-7b,
    \item LLaMA-13b \citep{llama},
    \item LLaMA-2-7b,
    \item LLaMA-2-13b \citep{llama2},
    \item Mistral-7b \cite{mistral}.
\end{enumerate*} We evaluate LLM quality using the perplexity metric on 2 datasets: WikiText-2 \citep{wikitext2} and C4 \citep{c4}, and accuracy on 3 benchmarks in zero-shot setting: WinoGrande \citep{winogrande}, PIQA \citep{piqa}, and ARC Challenge \citep{arc}. Perplexity is evaluated on the test set of the datasets at the maximum context length of the LLM (2048 for LLaMA, 4096 for LLaMA-2, and 8192 for Mistral).

\textbf{Baselines} We compare our proposed approach with uncompressed FP16 KV cache and competitive KV cache quantization methods, including \begin{enumerate*}
    \item uniform integer (INT) quantization (without grouping and with a group size of 128),
    \item NormalFloat (NF) quantization \citep{qlora} (without grouping and with a group size of 128),
    \item KVQuant \citep{kvquant} (without sparse outliers and with 1\% outliers stored in sparse format).
\end{enumerate*} KVQuant-\texttt{<b>}b-1\% is a dense-and-sparse method that requires an additional sparse matrix multiplication in addition to the dense matrix multiplication during inference, which introduces additional computational overhead. KVQuant and our proposed CQ both require learning centroids on a calibration dataset, and we use the same calibration set of 16 sequences (each with 2048 tokens) of WikiText-2 for both methods. Calibration is performed once on the training set of WikiText-2, while perplexity and accuracy are evaluated on the test sets of different datasets and benchmarks. For each method, we report bits per floating-point number (FPN) to measure the compression rate, which is calculated as the number of bits in the quantized KV cache of each token divided by the number of FPN in the uncompressed KV cache of each token, excluding the constant storage overheads of centroids, scaling factors, and zero points.

\begin{table}[t]
\caption{Perplexity on C4 under different KV cache quantization methods at varying bit-width. The results of INT, NF, and KVQuant (excluding 1b and 1b-1\%) are from \citep{kvquant}. Our method CQ consistently outperforms non-dense-and-sparse quantization methods, and performs better or on par with the dense-and-sparse method KVQuant-\texttt{<b>}b-1\%.}
\setlength\tabcolsep{4pt}
\centering
\small
\label{tab:ppl_c4}
\begin{tabular}{lr|ccccc}
\toprule
 & Bits Per FPN & LLaMA-7b & LLaMA-13b & LLama-2-7b & LLaMA-2-13b & Mistral-7b \\
 \midrule
 \midrule
FP16 & 16 & 7.08 & 6.61 & 6.63 & 6.05 & 5.71 \\
 \midrule
INT4 & 4.00-4.01 & 7.40 & 6.82 & 7.31 & 6.59 & 5.91 \\
INT4-gs128 & 4.16 & 7.16 & 6.67 & 6.87 & 6.20 & 5.76 \\
NF4 & 4.00-4.01 & 7.27 & 6.74 & 7.09 & 6.45 & 5.85 \\
NF4-gs128 & 4.16 & 7.16 & 6.66 & 6.86 & 6.20 & 5.77 \\
KVQuant-4b & 4.00-4.02 & 7.13 & 6.65 & 6.70 & 6.11 & 5.75 \\
KVQuant-4b-1\% & 4.32-4.35 & \textbf{7.09} & \textbf{6.62} & \textbf{6.65} & \textbf{6.06} & \textbf{5.72} \\
CQ-2c8b & 4.00 & \underline{7.11} & \underline{6.64} & \underline{6.67} & \underline{6.09} & \underline{5.74}\\
 \midrule
INT2 & 2.00-2.01 & 10892 & 100870 & 4708 & 4220 & 477 \\
INT2-gs128 & 2.14 & 43.49 & 56.25 & 113.49 & 97.04 & 50.73 \\
NF2 & 2.00-2.02 & 2850.1 & 4680.3 & 13081.2 & 4175.6 & 1102.3 \\
NF2-gs128 & 2.14 & 248.32 & 118.18 & 420.05 & 499.82 & 191.73 \\
KVQuant-2b & 2.00-2.02 & 10.28 & 9.05 & 15.16 & 43.77 & 8.40 \\
KVQuant-2b-1\% & 2.32-2.35 & \textbf{7.38} & \textbf{6.83} & \textbf{7.06} & \textbf{6.38} & \textbf{6.08} \\
CQ-4c8b & 2.00 & \underline{7.52} & \underline{6.96} & \underline{7.23} & \underline{6.52} & \underline{6.17} \\
 \midrule
KVQuant-1b & 1.00-1.02 & 168.90 & 1316.41 & 362.94 & 4223.37 & 127.07 \\
KVQuant-1b-1\% & 1.32-1.35 & \underline{11.18} & \underline{9.56} & 16.04 & 22.87 & 10.53 \\
CQ-8c8b & 1.00 & 12.13 & 10.53 & \underline{12.49} & \underline{10.53} & \underline{9.89} \\
CQ-8c10b & 1.25 & \textbf{9.12} & \textbf{8.23} & \textbf{9.03} & \textbf{8.01} & \textbf{7.46} \\
\bottomrule
\end{tabular}
\end{table}

\subsection{Results}

The results of perplexity on WikiText-2 are presented in Table \ref{tab:ppl_wiki} and the results on C4 are presented in Table \ref{tab:ppl_c4}. CQ consistently outperforms non-dense-and-sparse quantization methods, especially in low bit-width regions. Despite using lower bit-width, CQ performs better or on par with the dense-and-sparse quantization method KVQuant-\texttt{<b>}b-1\%. Dense-and-sparse quantization methods introduce additional inference overhead due to the extra sparse matrix multiplications for activation outliers which is inefficient on GPUs, while CQ does not have this limitation.

The accuracy results on different benchmarks are shown in Table \ref{tab:accuracy}. CQ consistently outperforms the non-dense-and-sparse baseline KVQuant-\texttt{<b>}b at bit-width of 1 and 2, and performs better or on par with the dense-and-sparse baseline KVQuant-\texttt{<b>}b-1\%.

\begin{table}[t]
\caption{Accuracy on 3 benchmarks under different KV cache quantization methods at varying bit-width.}
\setlength\tabcolsep{4.5pt}
\centering
\scriptsize
\label{tab:accuracy}
\begin{tabular}{lr|r|ccccc}
\toprule
 & Bits Per FPN & Task & LLaMA-7b & LLaMA-13b & LLaMA-2-7b & LLaMA-2-13b & Mistral-7b \\
 \midrule
 \midrule
\multirow{3}{*}{FP16} & \multirow{3}{*}{16} & WinoGrande & 69.93 & 72.69 & 68.90 & 71.98 & 73.88 \\
 &  & PIQA & 78.67 & 79.16 & 78.07 & 79.16 & 80.58 \\
 &  & ARC Challenge & 41.72 & 46.42 & 43.43 & 48.29 & 50.34 \\
 \midrule
\multirow{3}{*}{KVQuant-4b} & \multirow{3}{*}{4.00-4.02} & WinoGrande & 69.53 & \underline{72.61} & 67.96 & 71.59 & \textbf{73.88} \\
 &  & PIQA & \textbf{78.62} & \textbf{79.22} & 77.86 & \underline{78.94} & \underline{80.58} \\
 &  & ARC Challenge & \textbf{42.32} & \underline{45.99} & 42.75 & 46.67 & 49.06 \\
 \greyrule
\multirow{3}{*}{KVQuant-4b-1\%} & \multirow{3}{*}{4.32-4.35} & WinoGrande & \textbf{70.72} & \textbf{73.40} & \textbf{68.67} & \underline{72.30} & \underline{73.72} \\
 &  & PIQA & 78.40 & \underline{79.16} & \textbf{78.07} & \textbf{79.27} & \textbf{80.74} \\
 &  & ARC Challenge & 41.38 & \textbf{46.76} & 43.17 & \textbf{47.87} & \textbf{49.91} \\
 \greyrule
\multirow{3}{*}{CQ-2c8b} & \multirow{3}{*}{4.00} & WinoGrande & \underline{70.40} & 72.45 & \underline{68.27} & \textbf{72.53} & 73.48 \\
 &  & PIQA & \underline{78.61} & 79.11 & \underline{77.91} & 78.62 & 80.52 \\
 &  & ARC Challenge & \underline{41.55} & \underline{45.99} & \textbf{43.34} & \underline{47.78} & \underline{49.15} \\
 \midrule
\multirow{3}{*}{KVQuant-2b} & \multirow{3}{*}{2.00-2.02} & WinoGrande & 53.59 & 59.35 & 51.70 & 51.30 & 63.46 \\
 &  & PIQA & 72.47 & 74.81 & 63.38 & 65.40 & 75.46 \\
 &  & ARC Challenge & 32.00 & 34.47 & 22.44 & 24.66 & 38.57 \\
 \greyrule
\multirow{3}{*}{KVQuant-2b-1\%} & \multirow{3}{*}{2.32-2.35} & WinoGrande & \textbf{68.03} & \textbf{71.43} & \textbf{67.64} & \textbf{70.17} & \textbf{70.80} \\
 &  & PIQA & \textbf{77.69} & \textbf{78.51} & \textbf{76.60} & \textbf{78.51} & \textbf{79.65} \\
 &  & ARC Challenge & \textbf{38.74} & \textbf{45.14} & \textbf{41.47} & \textbf{44.97} & \textbf{47.53} \\
 \greyrule
\multirow{3}{*}{CQ-4c8b} & \multirow{3}{*}{2.00} & WinoGrande & \underline{67.48} & \underline{70.72} & \underline{66.45} & \underline{69.06} & \underline{69.38} \\
 &  & PIQA & \underline{76.11} & \underline{78.29} & \underline{76.12} & \underline{77.42} & \underline{79.49} \\
 &  & ARC Challenge & \underline{38.48} & \underline{44.03} & \underline{39.93} & \underline{44.11} & \underline{45.65} \\
 \midrule
\multirow{3}{*}{KVQuant-1b} & \multirow{3}{*}{1.00-1.02} & WinoGrande & 50.51 & 48.46 & 50.91 & 49.41 & 49.80 \\
 &  & PIQA & 53.26 & 53.54 & 53.37 & 50.92 & 54.73 \\
 &  & ARC Challenge & 21.76 & 21.33 & 20.65 & 21.67 & 19.88 \\
 \greyrule
\multirow{3}{*}{KVQuant-1b-1\%} & \multirow{3}{*}{1.32-1.35} & WinoGrande & \underline{56.67} & 61.01 & \underline{57.77} & \underline{57.30} & 58.17 \\
 &  & PIQA & \underline{71.38} & \underline{75.46} & 69.91 & 70.89 & 73.83 \\
 &  & ARC Challenge & 29.69 & \underline{35.32} & \underline{31.48} & 32.59 & 33.19 \\
 \greyrule
\multirow{3}{*}{CQ-8c8b} & \multirow{3}{*}{1.00} & WinoGrande & 56.51 & \underline{61.56} & 55.01 & 57.14 & \underline{58.25} \\
 &  & PIQA & 71.16 & 73.99 & \underline{71.22} & \underline{73.01} & \underline{75.24} \\
 &  & ARC Challenge & \underline{30.20} & 33.79 & 30.20 & \underline{34.30} & \underline{33.79} \\
 \greyrule
 \multirow{3}{*}{CQ-8c10b} & \multirow{3}{*}{1.25} & WinoGrande & 	\textbf{60.46} & \textbf{65.27} & \textbf{59.19} & \textbf{62.98} & \textbf{63.93} \\
& & PIQA & \textbf{73.45} & \textbf{75.90} & \textbf{73.07} & \textbf{74.37} & \textbf{77.31} \\
 &  & ARC Challenge & \textbf{33.28} & \textbf{37.12} & \textbf{34.64} & \textbf{38.74} & \textbf{39.59} \\
 \bottomrule
\end{tabular}
\end{table}

\subsection{Ablation Study}

We perform a set of ablation experiments to study the effectiveness of each component of our proposed approach. The results of the ablation experiments are shown in Table \ref{tab:ablation}. We evaluate the perplexity of 2 models Mistral-7b and LLaMA-2-13b on WikiText-2 using CQ at 2 bits per FPN, with varying number of channels coupled and comparing uniform centroid learning and Fisher-guided centroid learning. Fisher-guided centroids significantly improve model quality as demonstrated by lower perplexity. With either uniform centroids or Fisher-guided centroids, perplexity improves as the number of coupled channels increases. Hence, our proposed techniques of channel coupling and Fisher-guided centroid learning are both effective for maintaining model quality.

\begin{table}[t]
\caption{Ablation study: perplexity on WikiText-2 using CQ with varying number of coupled channels and fisher-guide centroids. Perplexity consistently improves as the number of coupled channels increases.}
\setlength\tabcolsep{5pt}
\centering
\scriptsize
\label{tab:ablation}
\begin{tabular}{l|cccccc|cccccc}
\toprule
                         & \multicolumn{6}{c|}{Mistral-7b}            & \multicolumn{6}{c}{LLaMA-2-13b}             \\
\midrule
Bits Per FPN             & 2     & 2     & 2    & 2    & 2    & 2    & 2      & 2      & 2    & 2    & 2    & 2    \\
Num. of Channels Coupled & 1     & 2     & 4    & 1    & 2    & 4    & 1      & 2      & 4    & 1    & 2    & 4    \\
Fisher-guided Centroids  & \xmark    & \xmark    & \xmark   & \cmark  & \cmark  & \cmark  & \xmark     & \xmark     & \xmark   & \cmark  & \cmark  & \cmark  \\
\midrule
Perplexity $\downarrow$               & 97.76 & 16.29 & 5.42 & 5.34 & 5.20 & 5.11 & 890.42 & 171.96 & 6.62 & 6.06 & 4.91 & 4.81 \\
\bottomrule
\end{tabular}
\end{table}

\subsection{Overhead of Centroid Learning and Storage}
\label{sec:overhead}

In this section, we discuss the computational overhead of centroid learning and the memory overhead of centroid storage for CQ. The centroid learning process of CQ consists of many independent k-means runs, which can be time-consuming on CPUs. Hence, we leverage a GPU implementation to accelerate the learning process. In all our experiments, we use k-means++ initialization and run 100 iterations of k-means on a single GPU to obtain the centroids. The memory overhead of storing the centroids can be calculated as $l \times 2 \times h \times c \times 2^b$ 16-bit floating-point numbers, where $l$ is the number of layers, $2$ is for keys and values, $h$ is the number of key/value attention heads, $c$ is the number of channels in a single-head key/value activation embedding, and $b$ is the bit width of quantized codes. The detailed learning and memory overhead for different CQ configurations and models are given in Table \ref{tab:overhead}. CQ can easily scale to large model sizes with low learning and memory overheads.

\begin{table}[h]
\caption{Learning and memory overhead of different CQ configurations and models. The number of centroid parameters are shown in millions, and the percentage to the model weights is shown in brackets.}
\label{tab:overhead}
\setlength\tabcolsep{7pt}
\centering
\scriptsize
\begin{tabular}{l|ccc|ccc}
\toprule
           & \multicolumn{3}{c|}{Centroid Learning Time} & \multicolumn{3}{c}{Parameter Count in Centroids}        \\
          CQ Config. & 2c8b      & 4c8b      & 8c8b      & 2c8b             & 4c8b             & 8c8b             \\
\midrule
LLaMA-7b   & 53 mins   & 28 mins   & 14 mins   & 67.11M (0.996\%) & 67.11M (0.996\%) & 67.11M (0.996\%) \\
LLaMA-13b   & 94 mins & 44 mins & 22 mins & 104.86M (0.806\%) & 104.86M (0.806\%) & 104.86M (0.806\%) \\
LLaMA-2-7b  & 54 mins & 28 mins & 14 mins & 67.11M (0.996\%)  & 67.11M (0.996\%)  & 67.11M (0.996\%)  \\
LLaMA-2-13b & 83 mins & 44 mins & 23 mins & 104.86M (0.806\%) & 104.86M (0.806\%) & 104.86M (0.806\%) \\
Mistral-7b & 13 mins   & 7 mins    & 4 mins    & 16.78M (0.232\%) & 16.78M (0.232\%) & 16.78M (0.232\%) \\
\bottomrule
\end{tabular}
\end{table}

\section{Related Works}

The high memory requirements and computational demands of LLMs pose a great challenge to efficient inference. Post-training model weight quantization has been shown to be effective for reducing inference latency and memory requirements. GPTQ \citep{gptq} scales approximate Hessian-based weight quantization to large-scale models, and AWQ \citep{awq} preserves the salient weights in LLMs to achieve better quantized model quality. SqueezeLLM \citep{squeezellm} leverages sensitivity-based non-uniform clustering and dense-and-sparse quantization for preserving salient weights and outliers. In addition to weight quantization, KV cache compression approaches are also effective for improving inference efficiency. Scissorhands \citep{scissorhands} and H2O \citep{h2o} achieve KV cache compression while preserving model quality by evicting unimportant tokens and only storing pivotal tokens. KIVI \citep{kivi} quantizes key activations channel-wise and value activations token-wise, and uses a residual to achieve better model quality. KVQuant \cite{kvquant} proposes sensitivity-based quantization and dense-and-sparse quantization for KV cache. FlashAttention \citep{flashattention} improves the inference efficiency of attention-based models on GPUs by fusing kernels to eliminate unnecessary global memory reads/writes, while NoMAD-Attention \citep{nomad} accelerates attention computations on CPUs by leveraging in-register shuffles. Product Quantization \citep{pq} is an approximate nearest neighbor search method that compresses vectors by decomposing the vector space into a Cartesian product of low-dimensional subspaces, and jointly quantizing the dimensions within each subspace.
\section{Conclusion}

In this work, we propose Coupled Quantization (CQ) for enabling more efficient LLM inference by compressing KV cache, which is the latency and throughput bottleneck in long context or large batch size settings. We observe that distinct channels of key/value activation embeddings exhibit high levels of dependency, which has not been leveraged by existing compression methods. Motivated by this insight, we propose channel coupling for exploiting the inter-channel dependency to achieve more information-efficient encoding of key/value activations. Furthermore, we propose Fisher-guided centroid learning to better preserve salient activations and model quality. Extensive experiments demonstrate that our method mostly outperforms existing methods in terms of model quality under the same quantization bit width. Moreover, CQ can preserve model quality reasonably well with KV cache quantized down to 1-bit.

\bibliographystyle{plainnat}
\bibliography{ref}

\newpage
\appendix
\section*{Appendix}
\label{sec:corr}
\section{Correlation Matrices and Scatter Plots}

\begin{figure}[h]
  \centering
  \includegraphics[width=\textwidth]{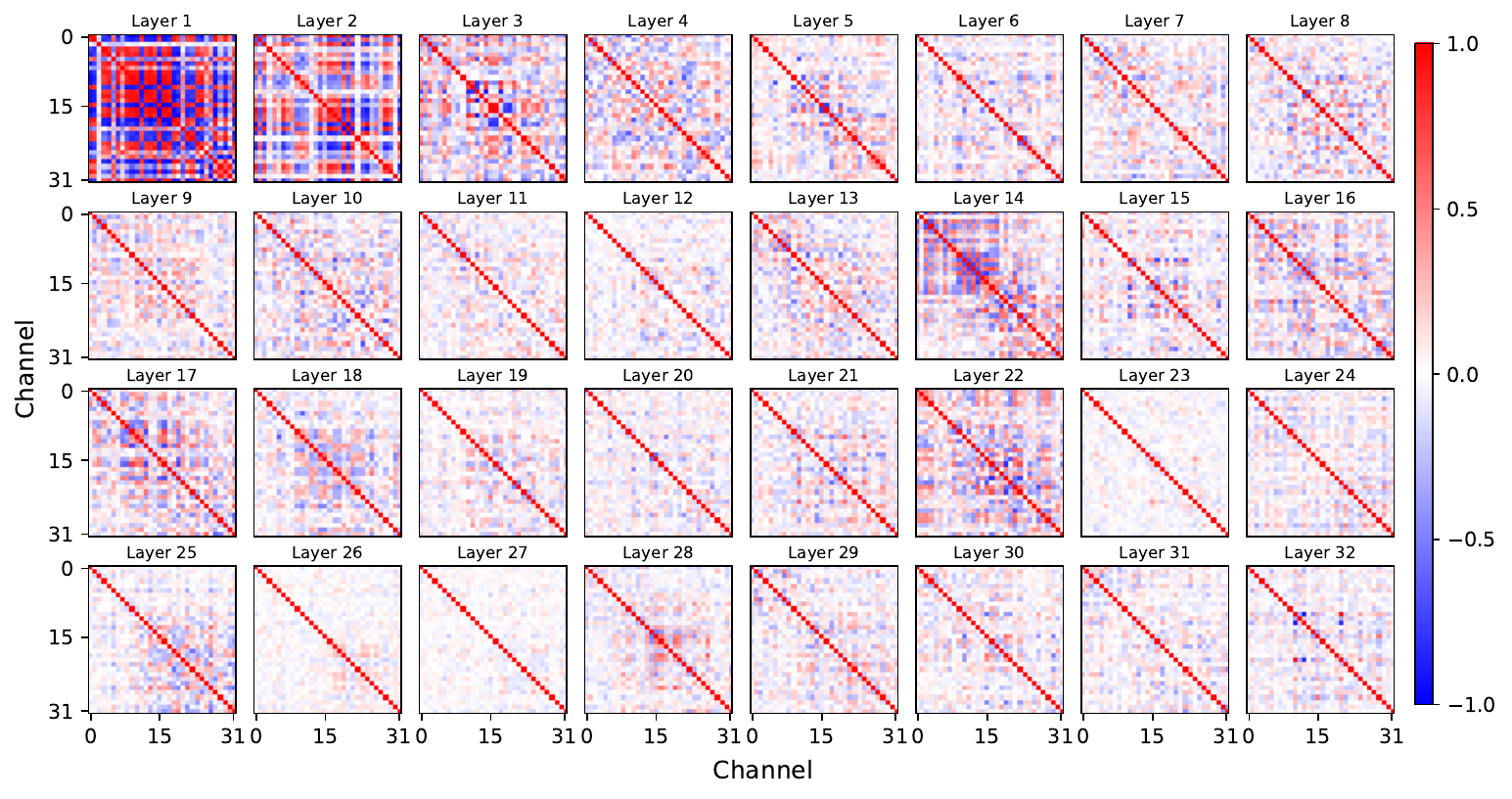}
  \caption{Correlation matrix for the first 32 channels of pre-RoPE \textbf{key} activation embeddings of all LLaMA-7b layers on WikiText-2.}
  \label{fig:key_corr}
\end{figure}

\begin{figure}[h]
  \centering
  \includegraphics[width=\textwidth]{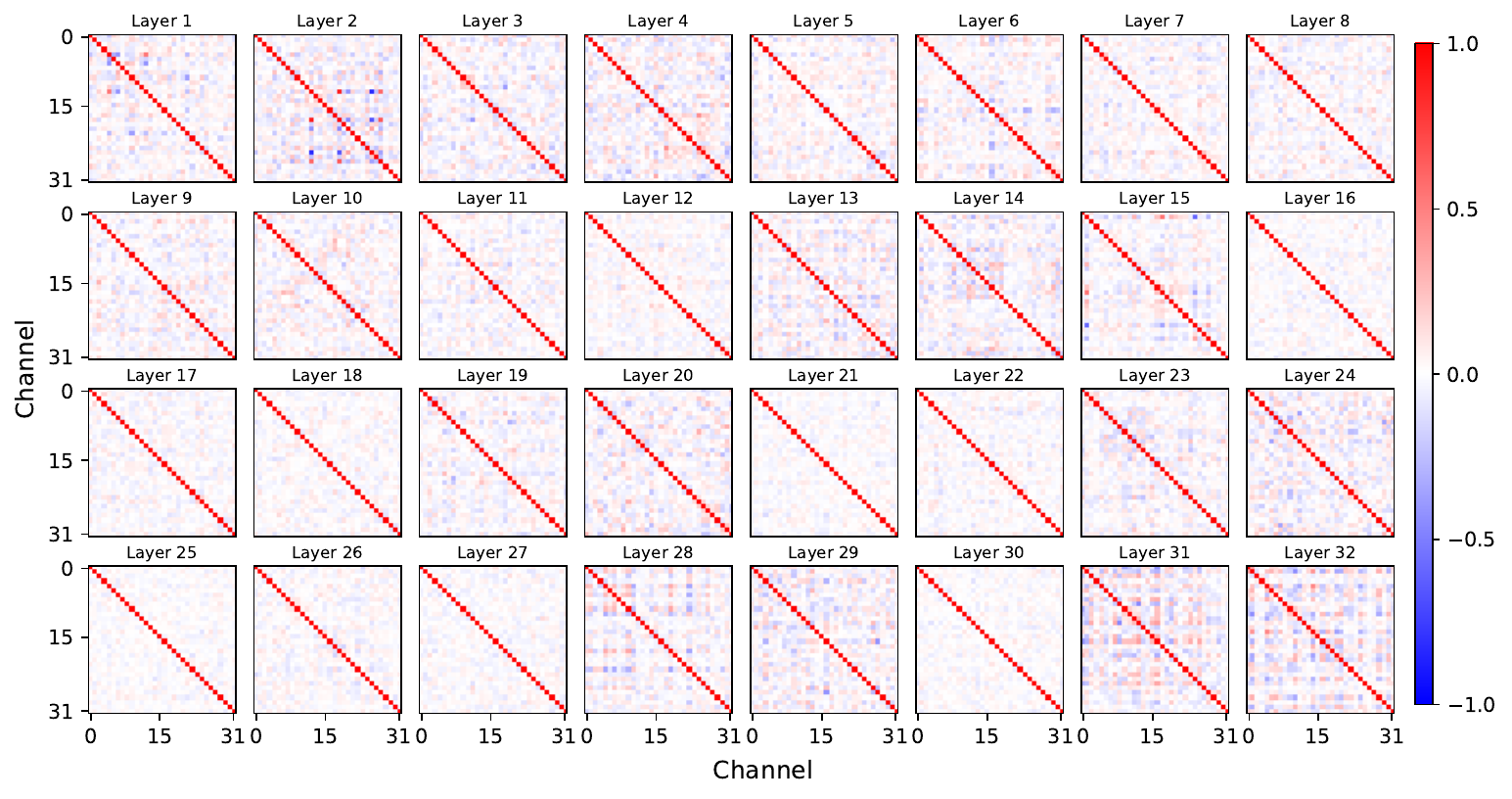}
  \caption{Correlation matrix for the first 32 channels of \textbf{value} activation embeddings of all LLaMA-7b layers on WikiText-2.}
  \label{fig:value_corr}
\end{figure}

\begin{figure}[h]
  \centering
  \includegraphics[width=\textwidth]{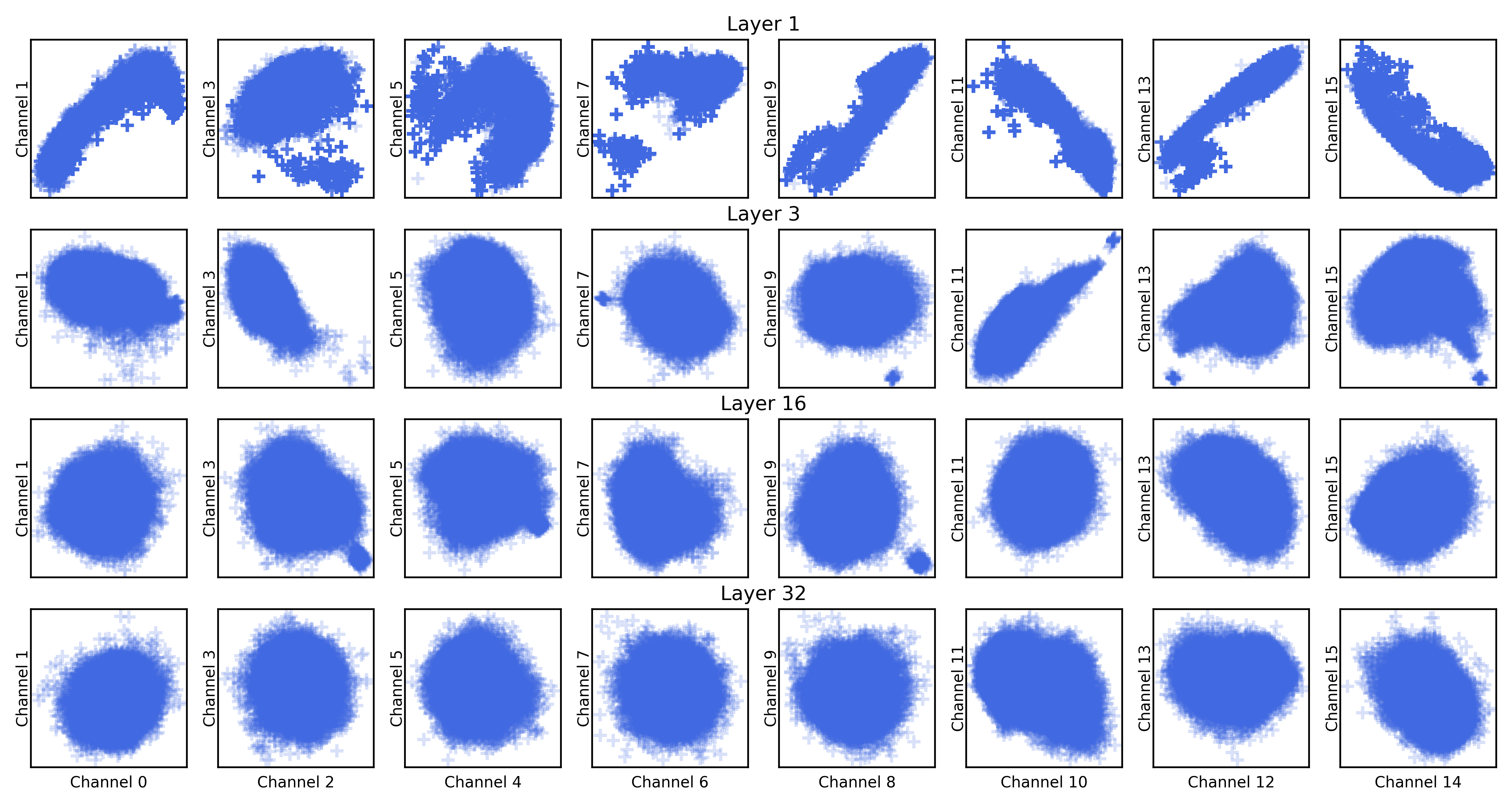}
  \caption{2-dimensional scatter plots of pairs of channels in \textbf{key} activation embeddings of 4 LLaMA-7b layers on WikiText-2.}
  \label{fig:key_emb}
\end{figure}

\begin{figure}[h]
  \centering
  \includegraphics[width=\textwidth]{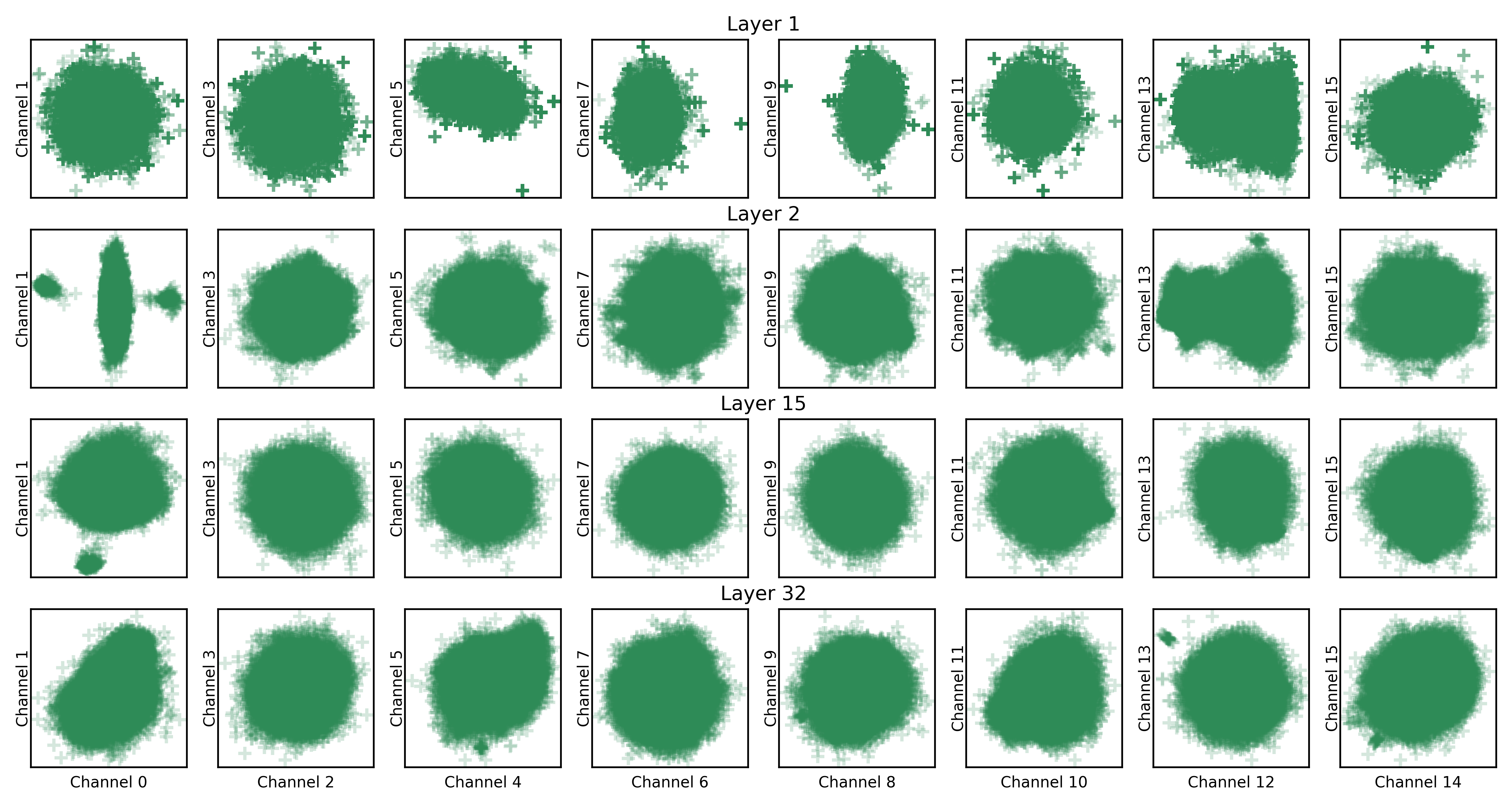}
  \caption{2-dimensional scatter plots of pairs of channels in \textbf{value} activation embeddings of 4 LLaMA-7b layers on WikiText-2.}
  \label{fig:value_emb}
\end{figure}

\end{document}